\documentclass[10pt,twocolumn]{article}

\usepackage[margin=0.72in,columnsep=0.24in]{geometry}
\usepackage{booktabs}
\usepackage{graphicx}
\usepackage{microtype}
\usepackage{tabularx}
\usepackage{enumitem}
\usepackage{xcolor}
\usepackage{placeins}
\usepackage{balance}
\usepackage{url}
\usepackage[hidelinks]{hyperref}
\usepackage{cite}
\newcolumntype{P}[1]{>{\raggedright\arraybackslash}p{#1}}
\newcolumntype{Y}{>{\raggedright\arraybackslash}X}
\hypersetup{
  pdftitle={Failure-Aware Long-Form Translation: Design and Implementation of a Recoverable LLM Translation System},
  pdfauthor={Yanlin Yu}
}

\newcommand{\sse}[1]{\texttt{#1}}

\title{Failure-Aware Long-Form Translation:\\
Design and Implementation of a Recoverable LLM Translation System}

\author{Yanlin Yu\\
Doctoral Program in International Public Policy\\
Degree Programs in Humanities and Social Sciences\\
University of Tsukuba, Japan}

\date{August 2026}

\begin{document}
\maketitle

\begin{abstract}
A long-form translation request can succeed at the API layer and still produce
an unusable result. The output may be empty, truncated, filtered, dominated by
source or prompt material, or interrupted after producing text worth keeping.
This report describes a recovery protocol developed for a deployed translation
system with heterogeneous inputs and provider APIs. It delays the first visible
release behind a 64-character window, validates the assembled output, and uses
typed stream events to distinguish replacement from continuation. Interrupted
work is retained only when a paragraph or sentence prefix can be re-derived
from the source. Further attempts follow a stable model order and a shared
deadline before entering a provenance-marked fallback path. A sanitized
companion artifact implements the protocol and passes 38 public tests. Its
fixed cases reproduce all 14 configured completion labels, contain four
early-invalid prefixes before any of their 235 characters become visible,
retain 31 boundary-safe characters across four interrupted streams, and satisfy
the attempt, event, and provenance rules in two end-to-end scenarios. These
results are executable checks of the published control flow. Translation
quality and detector performance on naturally occurring outputs require a
different evaluation.
\end{abstract}

\section{Introduction}

Consider a forty-part document for which thirty-five parts have already been
translated. During part thirty-six, an API continues to return a syntactically
valid stream, but the model begins reproducing the source or its hidden task
instructions. In another run, the connection ends after two complete
paragraphs and half of a sentence. A conventional failed-request abstraction
does not fully capture either event. The first may already have placed unusable or
sensitive internal text in the reader interface; the second may discard work
that was valid and expensive to produce.

This distinction matters because provider-level success is not
application-level success. A non-empty response can be truncated, contain
large source-language residue, or omit most of a multi-paragraph input. A
stream can terminate cleanly at the transport layer with a length or filtering
finish reason that makes the content unusable. Conversely, part of an
interrupted stream may be safe to retain. Long documents amplify the cost of
getting these distinctions wrong: full restart increases latency and cost,
while blindly retaining an unverified suffix can misalign subsequent text and
make omissions harder to detect.

Document-level machine translation research shows that context, document
position, and input length affect translation and its evaluation
\cite{vernikos2022documentmetrics,pal2024document,peng2025length}. Work on LLM
translation and literary refinement also motivates adequacy-sensitive
evaluation \cite{hendy2023gptmt,tan2026refinement}. This report addresses a
related application decision: whether text can be shown, retained, retried, or
attributed to a fallback while a request is running.

The implementation is a web-based system for manual input, authorized URL
acquisition, HTML and text files, PDFs, and OCR. The report concentrates on the
recovery path and makes three contributions:

\begin{enumerate}[leftmargin=*,nosep]
  \item a streaming architecture that distinguishes buffered, visible,
  boundary-safe, and committed text while recovery is still in progress;
  \item an operational failure matrix and bounded routing policy derived from
  the deployed implementation; and
  \item a sanitized executable artifact with public tests and fixed cases for
  release, retention, attempts, events, and provenance.
\end{enumerate}

The unit of evaluation is a recovery decision under a fixed runtime condition.
This choice keeps the evidence close to the implementation: the evaluation
cases exercise containment, retry, routing, and fallback, but do not compare
model quality or estimate failure prevalence.

\section{Requirements and System Context}

\subsection{Long-form and heterogeneous input}

The system accepts manual text and files in addition to source-specific URL
adapters. Acquisition operates within user authorization; restricted content
is supplied through a path the user is entitled to access. Parsing, OCR,
and normalization produce source
text and metadata; the text is then divided into request-sized chunks. A chunk
may contain several paragraphs, and a document contains multiple chunks.

This hierarchy shapes recovery behavior. The client tracks document-level completion by
chunk, while the server may recover within a chunk at complete paragraph or
sentence boundaries. Terms such as token, network delta, sentence, paragraph,
chunk, and committed document progress therefore need to remain distinct.

\subsection{Streaming and partial progress}

Waiting for an entire long document before displaying any result delays
visible progress and makes failure less transparent. The system therefore uses server-sent
events (SSE) for translation. Streaming introduces a release problem: once a
delta is appended to the interface, end-of-stream validation alone cannot
prevent its temporary exposure. It also introduces a recovery problem: retrying
the complete document after a late interruption wastes valid work, but treating
every received token as valid makes partial corruption persistent.

The design uses three distinct states. \emph{Buffered} text is not yet visible.
\emph{Visible} text has passed the leading guard but may still be replaced if a
later whole-output check fails. \emph{Committed} text has passed completion
checks and is recorded as a completed chunk. Within an interrupted attempt, a
\emph{boundary-safe prefix} contains only complete paragraphs and, conservatively,
complete sentence-like units that can be aligned to available source units.
The term describes a structural resume boundary rather than semantic
verification of the translation.

\subsection{Heterogeneous provider behavior}

Providers sharing an OpenAI-compatible interface still differ in streaming
support, timeout behavior, filtering, quota errors, terminal event structure,
and whether reasoning tokens precede visible content. A provider can also
return an empty message or an output that is syntactically well formed but
semantically not a translation. Accordingly, the orchestrator evaluates
application signals alongside exceptions and status codes.

\subsection{Operational requirements}

Table~\ref{tab:requirements} summarizes the design requirements. They arise
from finite serverless execution time, provider cost, the need for responsive
reading, and the requirement that recovery remain visible rather than silently
changing the origin of a translation.

\begin{table}[t]
\centering
\caption{Requirements and their implementation mechanisms.}
\label{tab:requirements}
\small
\begin{tabularx}{\columnwidth}{@{}P{0.25\columnwidth}YY@{}}
\toprule
Requirement & Failure if unmet & Mechanism \\
\midrule
Contain invalid prefixes & Source or prompt echo is exposed & Hold and inspect the leading window \\
Preserve completed work & Retry restarts the document & Chunk state and boundary-safe resume \\
Bound recovery & Cost or latency grows without limit & Shared deadline; each model at most once \\
Expose provenance & Machine fallback resembles LLM output & Typed SSE events and client labels \\
Reject silent omissions & Non-empty output masks missing content & Finish, residue, and omission checks \\
\bottomrule
\end{tabularx}
\end{table}

\section{System Overview}

\begin{figure*}[t]
  \centering
  \includegraphics[width=0.98\textwidth]{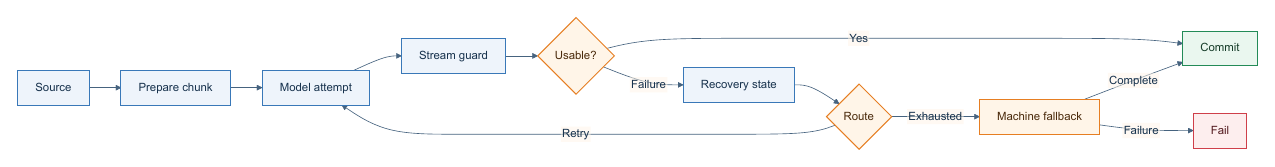}
  \caption{Translation and recovery flow. Every model result passes through the
  stream guard and completion decision. Failure derives a boundary-safe state, then
  returns to a bounded model route or an explicit machine-fallback path.}
  \label{fig:architecture}
\end{figure*}

Figure~\ref{fig:architecture} follows one request through translation and
recovery. Input adapters feed a common normalization and chunking layer.
Glossary information and the previous completed translation provide local
context. The session identifier also determines a stable order over the active
model pool. Chunks in the same task therefore do not reshuffle providers
independently, although the first choice can still vary across sessions.

The translation request is processed by an SSE handler. It emits metadata and
the current model, invokes a streaming model client, and passes candidate text
through a leading guard before emitting \sse{delta}. At completion it checks
finish reason, non-emptiness, prompt markers, source-language residue, and a
conservative large-omission heuristic. A valid result produces \sse{done} and
is committed as the current chunk.

Failures return to an explicit recovery policy. Before any delta is visible,
the system can rotate cleanly. After visibility, it either clears the current
attempt or retains only a re-derived paragraph/sentence prefix. Recovery uses a
finite model order and shared handler deadline. A separate compatibility path
may then try a model selected for filtering behavior, followed by an optional
configured external machine-translation service and a final emergency machine-translation
path. Machine fallbacks are non-streaming and must finish the entire remaining
chunk within budget; partial fallback output is rejected.

The frontend parses \sse{meta}, \sse{model}, \sse{reasoning}, \sse{delta},
\sse{restart}, \sse{replace}, \sse{done}, and \sse{error}. A \sse{restart}
announces a new attempt but does not change visible text. A \sse{replace}
resets the current chunk to a server-derived boundary-safe prefix. A \sse{delta} appends
text. The \sse{done} payload supplies the final combined text and a typed
fallback origin. This separation makes model switching observable and prevents
the client from guessing whether to append or overwrite.

\section{Operational Failure Matrix}

Table~\ref{tab:failure-matrix} collects failure conditions handled during
implementation and regression work. A single incident may match several rows.
The table is therefore a checklist for detection and recovery, not a count of
how often each failure occurs. Its coverage follows the provider and language
paths exercised by this implementation.

\begin{table*}[t]
\centering
\caption{Deployment-derived operational failure matrix. Detection combines
protocol checks with implementation-specific heuristics; rows may overlap.}
\label{tab:failure-matrix}
\small
\begin{tabularx}{\textwidth}{@{}P{0.14\textwidth}P{0.20\textwidth}YP{0.20\textwidth}@{}}
\toprule
Failure layer & Observable symptom & Detection and action & Residual risk \\
\midrule
Transport / invocation & Connection error, idle stream, or unsupported streaming & Provider iterator or deadline: rotate before release; otherwise replace with a boundary-safe prefix and resume & Blocking sockets or provider-specific errors \\
Provider or account & Authentication, rate, quota, or billing error & Exception classifier: mark an exhausted model and rotate & Evolving wording and status codes \\
Policy or filter & Filtering finish or classified refusal & Completion checks: discard the attempt and enter the compatibility or fallback path & Implicit refusals may resemble ordinary text \\
Terminal completion & Empty output or length finish & Terminal event: reject and rotate; never commit partial output & Missing or inaccurate finish metadata \\
Output adequacy & Source return or large source-language residue & Prefix and full-output checks: hold or replace, then rotate & Names, code mixing, and short inputs affect heuristics \\
Prompt boundary & Task, rule, source, or constraint markers & Leading and final checks: reject before release or replace afterward & Paraphrased leakage lacks known markers \\
Output coverage & Low paragraph and character retention & Completion check and client backstop: reject and retranslate the chunk & Thresholds may miss subtle omissions \\
Document continuity & Broken continuation, terminology drift, or an unverifiable prefix & Context and resume checks: add prior text or glossary; reject an invalid checkpoint & No semantic consistency guarantee \\
Provenance or persistence & Missing fallback type or source identity mismatch & Completion and version checks: type fallbacks and bind versions to sources & Legacy records and manual edits need migration \\
\bottomrule
\end{tabularx}
\end{table*}

The matrix also distinguishes detection confidence from action safety. For
example, an early prompt marker is a high-precision signal for this prompt
format, while source-language residue is language-dependent. The latter is
therefore tuned conservatively. English residue requires a sufficiently long
run rather than any Latin name; Japanese uses kana rather than all Han
characters; Korean uses Hangul density for severe-prefix detection. These
choices reduce obvious false positives but cannot establish semantic
completeness.

The omission guard addresses a different failure: a model may return a fluent,
non-empty ending while silently dropping most input paragraphs. The current
implementation applies it only to sources with at least four content blocks and
400 visible characters. It rejects output only when paragraph retention is
below 0.70 and the output-to-source character ratio is below 0.15. These are
incident-derived engineering parameters rather than general adequacy
thresholds. An output rejected
for omission is excluded from positional resume because a surviving final
paragraph could otherwise be aligned to the beginning of the source.

\section{Guarded Streaming and Containment}

\subsection{Leading-window guard}

The first containment mechanism operates before content becomes visible. The
streaming translator initially accumulates content instead of immediately
calling the client-visible delta callback. At the recorded snapshot, the hold
window is 64 characters. Each arriving fragment extends the buffered prefix.
If a known prompt marker appears, the attempt raises an unusable-output error
before any candidate content is emitted. Once the window is large enough, a
source-heavy check runs; failure similarly prevents release. If checks pass,
the accumulated prefix is flushed once and subsequent deltas flow normally.
A short valid translation that ends before filling the window is checked and
flushed at completion.

The window sets an explicit tradeoff between early containment and time to first
visible text. A larger window contains longer delayed echoes and increases
release latency. Markers can also appear after release, so completion repeats
the full-output check. A late failure triggers \sse{replace} and removes the
current attempt from view.

\subsection{Completion validity}

Building on the prefix guard, the system interprets provider completion
metadata as an application signal.
A content-filter finish enters the policy-specific path; a length finish is an
unusable truncation; and a successful terminal event with no content is an
empty-output failure. After assembly, prompt markers, severe source residue,
and catastrophic omission are checked again. Machine-translation fallbacks
translate bounded subchunks but return a result only if all subchunks complete
before the deadline. This all-or-nothing rule prevents a fallback from
converting a timeout into a plausible-looking half translation.

\subsection{State transitions}

\begin{figure}[t]
  \centering
  \includegraphics[width=0.95\columnwidth]{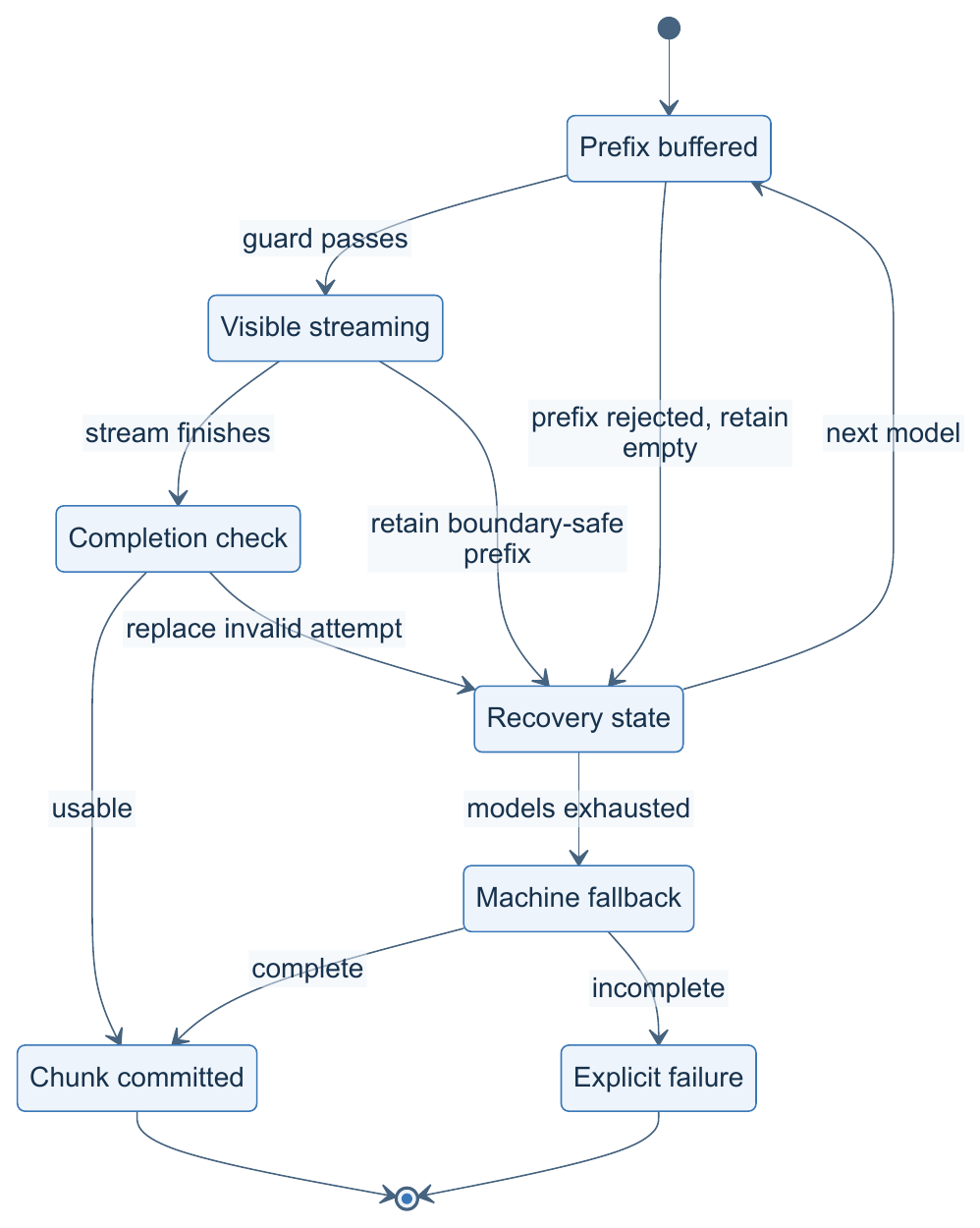}
  \caption{Guarded stream state machine. Every attempt begins with prefix
  buffering. An invalid or interrupted attempt moves through boundary-safe
  recovery before another model or machine fallback runs.}
  \label{fig:state}
\end{figure}

Figure~\ref{fig:state} shows the stream state machine. The key distinction is
between a failure before visible release and an interruption after release. In
the first case, rotation requires no text repair. In the second, the server
derives a safe state from the source and received output. It keeps completed
paragraphs. For the latest paragraph, it may also retain complete
sentence-like units ending in punctuation, bounded by the number of complete
source sentences. It drops the unterminated output fragment and reconstructs
the remaining source.

The server sends \sse{replace} with the retained text and the separator needed
to join the next output. The next model receives the remaining source as a new
translation unit and the boundary-safe prefix as previous context. Progress resets
the consecutive-failure counter; attempts that make no boundary-safe progress
consume the retry budget. A client-provided reconnect checkpoint is accepted
only if the server can re-derive exactly the same boundary-safe prefix and a non-empty
remainder from the original chunk.

Within these constraints, the mechanism preserves useful work under
non-atomic generation. Sentence
boundaries are punctuation heuristics, paragraph alignment is positional, and
generation has no exactly-once guarantee. The resulting release invariant is
deliberately narrow:
structurally incomplete trailing fragments and outputs already classified as unusable stay
outside resume context.

\section{Routing, Fallback, and Progress Preservation}

\subsection{Stable order and error-aware rotation}

Containment alone does not determine what happens after a failed attempt. For
a model tier and session identifier, the server deterministically shuffles
active models. Every chunk in the task therefore sees a stable order unless
operational state changes. Without a session identifier, the order begins at a
persisted current index. Quota-classified errors mark a model temporarily
exhausted; unusable outputs rotate without globally marking the model, because
the failure may be input-specific. Other provider errors are surfaced to the
handler rather than hidden behind an unbounded library retry. The underlying
client is configured with no automatic retries in the recorded deployment.

During stream recovery, the remaining source is treated as a new bounded
attempt. Models known not to support stable streaming are skipped for that
request. A compatibility path excludes models already attempted in the main
pool and tries each eligible model at most once. This fixed operational policy
exposes its model order and retry bound directly.

At the recorded snapshot, the handler budget is 240 seconds and the streaming
phase receives 170 seconds of that budget. Six consecutive failures without a
new boundary-safe prefix end stream recovery; retained progress resets the
counter because the remaining source has changed. These values are deployment
parameters and are reproduced in the companion artifact.

\subsection{Layered fallback}

After bounded model rotation is exhausted or its deadline is reached, the handler
preserves any boundary-safe prefix and applies a layered fallback to the remaining
source. The order is: a still-eligible compatibility model, an optional
configured external machine-translation provider, and an emergency machine-translation
path. Each layer receives only a bounded portion of the shared handler budget,
with time reserved for the last path. The recorded caps are 35 seconds for the
compatibility layer and 20 seconds for the external machine-translation layer,
with 10 seconds reserved for emergency machine translation. Catastrophic omission skips another LLM
compatibility attempt and retranslates the complete chunk, because the rejected
output is not a boundary-safe prefix.

Machine fallback is marked in the \sse{done} payload. The client keeps
model-compatible fallback separate from machine fallback and can display the
corresponding source comparison or retry affordance. Fallback provenance is
part of correctness: a complete machine translation is operationally preferable
to a silent gap, while the interface should continue to distinguish it from the
requested LLM path.

\subsection{Document-level progress}

The same recovery policy also operates at document level. The client holds an
array of source chunks and translated parts. A chunk is
initially marked interrupted, removed from that set only after non-empty
completion, and retained as retryable if the request fails or a client-side
omission backstop fires. Completed chunks remain in the assembled output while
a later chunk is retried. Translation versions bind text to source identity and
provenance metadata; these product-level mechanisms are not required for the
stream guard, but they prevent recovery state from being confused with a
different document revision.

\section{Implementation and Offline Evaluation}

\subsection{Deployed implementation and reference artifact}

The deployed implementation uses a Python serverless API and a browser client
written in vanilla JavaScript, with TypeScript used for static checking. The
paper reports the recovery protocol rather than the surrounding product. A
sanitized, provider-neutral companion artifact is included as an ancillary
archive with the submission. It implements the leading guard, completion
checks, boundary-safe retention, reconnect validation, typed stream events,
attempt bounds, shared budgets, layered fallback, all-or-nothing subchunk
assembly, and provenance markers described in Sections 5 and 6.

The artifact excludes the production product, user data, acquisition adapters,
accounts, community features, database code, credentials, and deployment
configuration. It uses only the Python standard library. Its 38 public tests
pass in an isolated run and cover guard boundaries, multilingual residue,
omission thresholds, resume-state derivation, reconnect validation, event
transitions, attempt bounds, deadlines, layered fallback, and provenance. The
larger private product regression suite is retained as implementation history,
but it is not used as independently reproducible evidence in this report.

\subsection{Fixed offline conformance set}

The companion artifact also contains fixed, hand-authored cases for four
mechanisms. The set combines valid near-boundary cases with injected prompt
echo, source return, terminal failure, omission, interruption, attempt
exhaustion, and fallback conditions. Table~\ref{tab:offline} reports the
results produced by the artifact's evaluation command.

\begin{table*}[t]
\centering
\caption{Offline conformance results from the public companion artifact.}
\label{tab:offline}
\small
\begin{tabularx}{\textwidth}{@{}P{0.18\textwidth}P{0.10\textwidth}P{0.25\textwidth}Y@{}}
\toprule
Mechanism & Cases & Comparison & Recorded result \\
\midrule
Completion guard & 14 & Configured usable/unusable labels & 0 false acceptances; 0 false rejections \\
Leading window & 4 early-invalid prefixes & Immediate release & 0 guarded visible characters versus 235 without the hold window \\
Interruption recovery & 4 interrupted outputs & Full restart & 4/4 expected boundary matches; 31 characters retained versus 0 \\
Bounded protocol & 2 end-to-end scenarios & Configured attempt, event, and provenance invariants & 0 attempt-bound, resume-event, or provenance violations \\
\bottomrule
\end{tabularx}
\end{table*}

Across the fixed set, every configured label and each scripted attempt, event,
and provenance invariant was reproduced. Table~\ref{tab:offline} reports the
leading-window and interruption comparisons.

This is internal conformance evidence. The cases were written to exercise rule
boundaries and use neither live provider calls nor production traces. They
cannot estimate detector recall, false-positive rates, translation quality, or
failure prevalence outside the supplied cases.

\FloatBarrier
\section{Discussion and Design Lessons}

\subsection{How the protocol changed}

An early resume path tried to preserve complete sentences whenever a stream was
interrupted. In practice, a sentence could be retained without a reliable
paragraph anchor and then aligned to the wrong part of the source. The visible
translation looked continuous while a source sentence was silently skipped.
Sentence rescue was consequently limited to cases with an established
paragraph prefix and disabled for moderation-related interruptions, where the
interrupted output did not provide a trustworthy continuation boundary. A
later frontend defect exposed the other side of the same problem: treating
every \sse{restart} as an instruction to clear the chunk erased valid streamed
text. The event was changed to an announcement of a new attempt, while only a
server-derived \sse{replace} could alter the retained prefix.

Prompt echo was initially checked only after the stream ended. In one recorded
response, a provider returned the task and source tags themselves, so the
interface displayed them before the final validator rejected the response. In
the same internal trace, holding the first 64 characters prevented that
exposure and delayed first visible text by about 1.5 seconds. The chosen window
is a deployment compromise, not a claim that 64 characters is optimal for
other prompts or interfaces.

In another internal incident, a 30-paragraph chunk returned one fluent
paragraph and passed every existing non-emptiness, prompt-echo, and residue
check. Because the client treated non-empty completion as success, repeated
manual retries were marked complete. The saved document then contained 600
translated paragraphs for a source document with 629 paragraphs. The paired
paragraph-retention and character-ratio test was added in response. Recovery
after catastrophic omission also changed: the rejected output no longer
supplies a resume prefix, because its surviving paragraph may not occupy the
same position in the source. The complete chunk is translated again by the
bounded machine path if model rotation cannot recover it.

Two fixed compatibility models were initially introduced to make
filtering-related retries predictable. Over time, that path accumulated
duplicate attempts and references to retired models. In one internal replay of
the same recorded failure sequence, the old path issued 104 LLM request
attempts, compared with 32 after inactive and already attempted models were
excluded. The present at-most-once rule replaced the fixed pair because it made
the bound depend on the active pool rather than on an aging list of model
names.

\subsection{What the evaluation establishes}

Transport completion and application-level acceptance are separate events in
this system. HTTP status and stream closure establish the former. Finish
metadata, visible text, and the remaining source still have to agree before a
chunk is committed. In the fixed cases, the leading window contained all 235
characters belonging to four known-invalid prefixes. A final validator would
have rejected the same outputs, but only after those prefixes had appeared in
the interface.

Interruption recovery needs a state between keeping every token and restarting
the document. The boundary-safe prefix serves that role. Across four fixed
interruptions, it retained 31 characters that a full restart discarded. This
amount is case-specific, but the case exercises the transition that the earlier
sentence-alignment failure made necessary. Semantic continuity remains an open
question because the alignment rule is structural and always removes the
unfinished suffix.

The failure matrix is best read in this operational frame. It records signals
that changed release, replacement, retry, or fallback behavior during
implementation. Some signals are precise for the current prompt and event
format; others are conservative heuristics. Separating them makes later
comparisons possible without turning this development history into a frequency
estimate.

\section{Related Work}

\subsection{MT evaluation, quality estimation, and output pathologies}

BLEU established inexpensive reference-based corpus evaluation
\cite{papineni2002bleu}; COMET uses multilingual representations trained on
human judgments \cite{rei2020comet}. Expert MQM work shows that
evaluation protocol and document context can materially affect conclusions
\cite{freitag2021experts}, while document-aware metric extensions explicitly
incorporate context \cite{vernikos2022documentmetrics}. Quality estimation
addresses a closer deployment setting by predicting word- or sentence-level
quality without a reference translation \cite{zerva2024qe}.

Hallucination and omission research is also directly relevant. HalOmi provides
human annotations for both phenomena across 18 translation directions
\cite{dale2023halomi}, while source-contribution and cross-lingual similarity
methods detect severe outputs detached from the source \cite{dale2023detecting}.
The guards in this report occupy a narrower operational role: configured
signals trigger release, replacement, retry, or fallback actions. Source
residue and omission checks overlap with adequacy assessment, but the current
evaluation measures the resulting control flow rather than detector accuracy
on naturally occurring translations.

Research on document-level corpora, length effects, GPT translation, and
literary refinement studies context use and output quality
\cite{pal2024document,peng2025length,hendy2023gptmt,tan2026refinement}. The
system uses chunking and previous-translation context. Its chunk size is an
implementation parameter whose quality tradeoffs remain to be measured.

\subsection{Uncertainty, abstention, and guarded output}

Semantic entropy detects confabulation by estimating uncertainty over meanings
in free-form generation \cite{farquhar2024semantic}. Broader abstention research
asks when a model should withhold an answer \cite{wen2025abstention}.
Agentic Abstention frames stopping as a sequential decision under an evolving
environment \cite{luo2026agentic}; AgentAbstain uses paired act/abstain tasks in
executable sandboxes \cite{liu2026agentabstain}; HiL-Bench evaluates whether
agents seek missing information instead of silently guessing
\cite{trinh2026hilbench}.

This literature addresses policy selection under uncertainty. The translation
orchestrator in this report uses a predefined bounded policy over observable
runtime signals. Learning when to accept, retry, route, fall back, or abstain
would provide a direct comparator, but this study does not train or evaluate
such a policy.

\subsection{Routing and fault-tolerant serving}

FrugalGPT studies learned LLM cascades for cost and performance
\cite{chen2023frugalgpt}; RouteLLM trains routers from preference data to trade
off stronger and weaker models \cite{ong2024routellm}. The stable model order
and fallback chain implement a fixed operational policy with unmeasured cost,
latency, and quality tradeoffs.
At a lower systems layer, D\'ej\`aVu preserves serving state such as KV caches
for fault-tolerant distributed inference \cite{strati2024dejavu}. Recovery in
this system assumes external provider APIs and preserves application-level
translation state. Infrastructure recovery and application validity are
complementary: restoring a generation process does not by itself determine
whether the recovered text is safe to release.

\section{Ethics, Privacy, and Responsible Deployment}

The recovery mechanisms sit within a broader set of content and deployment
responsibilities. Long-form translation can process copyrighted, private, sensitive, or
adult-oriented text. The system does not claim ownership of supplied content,
and users remain responsible for ensuring that acquisition and translation are
authorized. Source adapters are designed to preserve payment, age, and access
controls. When authorized URL
acquisition is unavailable, the user supplies text or a file they are entitled
to access.

The study uses synthetic test examples and isolated mock services. Production
databases, logs, user text, account information, private URLs, and credentials
remain outside the released evidence set. Private development records were
used only to verify the anonymized design history in Section 8; they are not
part of the public conformance evaluation. The paper withholds the product
name, address, and screenshots. The ancillary reference artifact contains the
guard, recovery, and conformance mechanisms and excludes acquisition,
authentication, payment, community, database, and deployment modules. A
pre-publication scan covers secrets, absolute paths, product identifiers, and
personal data.

Automatic translation can alter voice, identity, relationships, terminology,
or the representation of sensitive content. Output guards mitigate known
operational failures but do not ensure faithful translation. Machine fallback
is marked, and failed or residual-source chunks remain available for human
review and retry. These affordances reduce operational ambiguity, while
faithfulness and appropriate use still depend on human judgment. Research
using future production traces would require separate consent, minimization,
privacy review, and potentially institutional ethics review.

Generative AI language tools were used to assist with English-language
drafting, structural editing, and code organization. The author independently
designed the study, verified all claims, citations, experimental results, and
released code, edited the final manuscript, and assumes full responsibility for
its contents.

\section{Limitations}

This study is tied to one deployed system and its development history. The
matrix describes failures observed or anticipated in that system; it does not
measure their prevalence across products, providers, or domains. Several guards
are language-specific heuristics. Paraphrased prompt leakage, subtle omission,
and semantic inconsistency can pass them, while names, code mixing, and short
inputs can still produce false alarms. A boundary-safe prefix is structurally
aligned, not semantically certified.

The evidence comes from deterministic regression cases. Those cases are useful
for checking release, retry, event, and provenance behavior, but their counts
are correlated software tests rather than statistical samples. No systematic
live-provider evaluation is reported; isolated development observations are
not treated as evaluation results. There are no professional translation
labels or comparisons using BLEU, COMET, MQM, or human preference. The study
consequently says little about average translation quality or whether the
current routing order is optimal.

The deployment itself also constrains the design. Provider behavior, prices,
moderation policy, and latency change over time, while serverless deadlines
make bounded recovery especially important. A self-hosted stack would expose
different options. The original product mainly handles informal and literary
text, so transfer to legal, medical, technical, or speech translation remains
untested.

\section{Future Work}

A next experiment could compare recovery decisions rather than add more
scripted cases. One possible formulation is \emph{selective recovery}: given a
candidate output, runtime signals, and a remaining budget, choose whether to
accept, retry, route, fall back, or abstain. Paired decisions on independent
documents could then be judged against translation quality, latency, cost, and
release risk. Human assessment and explicit utility assumptions would be
needed, with analysis at document level rather than by correlated chunks.

\section{Conclusion}

Long-form translation does not fail only when a request crashes. It can also
fail when a response appears successful for the wrong reason: a non-empty
result may conceal missing paragraphs, expose prompt material, or end after
producing text that is worth keeping only at a safe boundary. These cases lead
to one design consequence: generating text cannot be the same as committing
it.

The reported system therefore treats release, recovery, and fallback as
separate decisions. It holds uncertain prefixes before they reach the reader,
re-derives the portion that can survive an interruption, and keeps the origin
of a fallback attached to the completed chunk. The fixed conformance cases make
this operational claim concrete. They show that the published protocol can
reject or preserve text without silently changing what the reader sees or where
the translation came from.

The accompanying artifact turns those decisions into executable checks.
Comparative translation quality, live-provider behavior, and the best recovery
policy remain empirical questions for a broader evaluation.

\balance
\bibliographystyle{plain}
{\small
\bibliography{references}

@inproceedings{papineni2002bleu,
  title     = {{BLEU}: a Method for Automatic Evaluation of Machine Translation},
  author    = {Papineni, Kishore and Roukos, Salim and Ward, Todd and Zhu, Wei-Jing},
  booktitle = {Proceedings of the 40th Annual Meeting of the Association for Computational Linguistics},
  pages     = {311--318},
  year      = {2002},
  doi       = {10.3115/1073083.1073135},
  url       = {https://aclanthology.org/P02-1040/}
}

@inproceedings{rei2020comet,
  title     = {{COMET}: A Neural Framework for {MT} Evaluation},
  author    = {Rei, Ricardo and Stewart, Craig and Farinha, Ana C. and Lavie, Alon},
  booktitle = {Proceedings of the 2020 Conference on Empirical Methods in Natural Language Processing},
  pages     = {2685--2702},
  year      = {2020},
  doi       = {10.18653/v1/2020.emnlp-main.213},
  url       = {https://aclanthology.org/2020.emnlp-main.213/}
}

@article{freitag2021experts,
  title   = {Experts, Errors, and Context: A Large-Scale Study of Human Evaluation for Machine Translation},
  author  = {Freitag, Markus and Foster, George and Grangier, David and Ratnakar, Viresh and Tan, Qijun and Macherey, Wolfgang},
  journal = {Transactions of the Association for Computational Linguistics},
  volume  = {9},
  pages   = {1460--1474},
  year    = {2021},
  doi     = {10.1162/tacl_a_00437},
  url     = {https://aclanthology.org/2021.tacl-1.87/}
}

@inproceedings{vernikos2022documentmetrics,
  title     = {Embarrassingly Easy Document-Level {MT} Metrics: How to Convert Any Pretrained Metric into a Document-Level Metric},
  author    = {Vernikos, Giorgos and Thompson, Brian and Mathur, Prashant and Federico, Marcello},
  booktitle = {Proceedings of the Seventh Conference on Machine Translation},
  pages     = {118--128},
  year      = {2022},
  doi       = {10.18653/v1/2022.wmt-1.6},
  url       = {https://aclanthology.org/2022.wmt-1.6/}
}

@inproceedings{zerva2024qe,
  title     = {Findings of the Quality Estimation Shared Task at {WMT} 2024: Are {LLM}s Closing the Gap in {QE}?},
  author    = {Zerva, Chrysoula and Blain, Frederic and C. De Souza, Jos{\'e} G. and Kanojia, Diptesh and Deoghare, Sourabh and Guerreiro, Nuno M. and Attanasio, Giuseppe and Rei, Ricardo and Orasan, Constantin and Negri, Matteo and Turchi, Marco and Chatterjee, Rajen and Bhattacharyya, Pushpak and Freitag, Markus and Martins, Andr{\'e}},
  booktitle = {Proceedings of the Ninth Conference on Machine Translation},
  pages     = {82--109},
  year      = {2024},
  doi       = {10.18653/v1/2024.wmt-1.3},
  url       = {https://aclanthology.org/2024.wmt-1.3/}
}

@inproceedings{dale2023detecting,
  title     = {Detecting and Mitigating Hallucinations in Machine Translation: Model Internal Workings Alone Do Well, Sentence Similarity Even Better},
  author    = {Dale, David and Voita, Elena and Barrault, Loic and Costa-juss{\`a}, Marta R.},
  booktitle = {Proceedings of the 61st Annual Meeting of the Association for Computational Linguistics (Volume 1: Long Papers)},
  pages     = {36--50},
  year      = {2023},
  doi       = {10.18653/v1/2023.acl-long.3},
  url       = {https://aclanthology.org/2023.acl-long.3/}
}

@inproceedings{dale2023halomi,
  title     = {{HalOmi}: A Manually Annotated Benchmark for Multilingual Hallucination and Omission Detection in Machine Translation},
  author    = {Dale, David and Voita, Elena and Lam, Janice and Hansanti, Prangthip and Ropers, Christophe and Kalbassi, Elahe and Gao, Cynthia and Barrault, Lo{\"i}c and Costa-juss{\`a}, Marta R.},
  booktitle = {Proceedings of the 2023 Conference on Empirical Methods in Natural Language Processing},
  pages     = {638--653},
  year      = {2023},
  doi       = {10.18653/v1/2023.emnlp-main.42},
  url       = {https://aclanthology.org/2023.emnlp-main.42/}
}

@inproceedings{pal2024document,
  title     = {Document-Level Machine Translation with Large-Scale Public Parallel Corpora},
  author    = {Pal, Proyag and Birch, Alexandra and Heafield, Kenneth},
  booktitle = {Proceedings of the 62nd Annual Meeting of the Association for Computational Linguistics (Volume 1: Long Papers)},
  year      = {2024},
  url       = {https://aclanthology.org/2024.acl-long.712/}
}

@inproceedings{peng2025length,
  title     = {Investigating Length Issues in Document-level Machine Translation},
  author    = {Peng, Ziqian and Bawden, Rachel and Yvon, Fran{\c{c}}ois},
  booktitle = {Proceedings of Machine Translation Summit XX: Volume 1},
  pages     = {4--23},
  year      = {2025},
  url       = {https://aclanthology.org/2025.mtsummit-1.3/}
}

@inproceedings{tan2026refinement,
  title     = {What Does {LLM} Refinement Actually Improve? A Systematic Study on Document-Level Literary Translation},
  author    = {Tan, Shaomu and Zhu, Dawei and Tran, Ke and Denkowski, Michael and Trenous, Sony and Ribeiro, Leonardo F. R. and Byrne, Bill and Hieber, Felix},
  booktitle = {Proceedings of the 64th Annual Meeting of the Association for Computational Linguistics (Volume 1: Long Papers)},
  pages     = {5929--5957},
  year      = {2026},
  doi       = {10.18653/v1/2026.acl-long.268},
  url       = {https://aclanthology.org/2026.acl-long.268/}
}

@article{hendy2023gptmt,
  title   = {How Good Are {GPT} Models at Machine Translation? A Comprehensive Evaluation},
  author  = {Hendy, Amr and Abdelrehim, Mohamed and Sharaf, Amr and Raunak, Vikas and Gabr, Mohamed and Matsushita, Hitokazu and Kim, Young Jin and Afify, Mohamed and Awadalla, Hany Hassan},
  journal = {arXiv preprint arXiv:2302.09210},
  year    = {2023},
  doi     = {10.48550/arXiv.2302.09210},
  url     = {https://arxiv.org/abs/2302.09210}
}

@article{farquhar2024semantic,
  title   = {Detecting Hallucinations in Large Language Models Using Semantic Entropy},
  author  = {Farquhar, Sebastian and Kossen, Jannik and Kuhn, Lorenz and Gal, Yarin},
  journal = {Nature},
  volume  = {630},
  pages   = {625--630},
  year    = {2024},
  doi     = {10.1038/s41586-024-07421-0},
  url     = {https://www.nature.com/articles/s41586-024-07421-0}
}

@article{wen2025abstention,
  title   = {Know Your Limits: A Survey of Abstention in Large Language Models},
  author  = {Wen, Bingbing and Yao, Jihan and Feng, Shangbin and Xu, Chenjun and Tsvetkov, Yulia and Howe, Bill and Wang, Lucy Lu},
  journal = {Transactions of the Association for Computational Linguistics},
  volume  = {13},
  pages   = {529--556},
  year    = {2025},
  doi     = {10.1162/tacl_a_00754},
  url     = {https://aclanthology.org/2025.tacl-1.26/}
}

@article{luo2026agentic,
  title   = {Agentic Abstention: Do Agents Know When to Stop Instead of Act?},
  author  = {Luo, Han and Wen, Bingbing and Wang, Lucy Lu},
  journal = {arXiv preprint arXiv:2606.28733},
  year    = {2026},
  url     = {https://arxiv.org/abs/2606.28733}
}

@article{liu2026agentabstain,
  title   = {{AgentAbstain}: Do {LLM} Agents Know When Not to Act?},
  author  = {Liu, Xun and Zhang, Yi Evie and Kasprova, Vira and Rabbani, Parisa and Zahraei, Pardis Sadat and Zhang, Tianyu and Ebrahimpour-Boroojeny, Ali and Chandrasekaran, Varun},
  journal = {arXiv preprint arXiv:2607.10059},
  year    = {2026},
  url     = {https://arxiv.org/abs/2607.10059}
}

@article{trinh2026hilbench,
  title   = {{HiL-Bench} (Human-in-Loop Benchmark): Do Agents Know When to Ask for Help?},
  author  = {Trinh, Tu and Elfeki, Mohamed and Luo, Guangze and Luu, Kelvin and Hunt, Nathan and Hern{\'a}ndez, Ernesto and Marwaha, Nandan and He, Yannis Yiming and Wang, Charles and Carabedo, Fernando and Castillo, Alessa and Liu, Bing},
  journal = {arXiv preprint arXiv:2604.09408},
  year    = {2026},
  url     = {https://arxiv.org/abs/2604.09408}
}

@article{chen2023frugalgpt,
  title   = {{FrugalGPT}: How to Use Large Language Models While Reducing Cost and Improving Performance},
  author  = {Chen, Lingjiao and Zaharia, Matei and Zou, James},
  journal = {arXiv preprint arXiv:2305.05176},
  year    = {2023},
  url     = {https://arxiv.org/abs/2305.05176}
}

@article{ong2024routellm,
  title   = {{RouteLLM}: Learning to Route {LLM}s with Preference Data},
  author  = {Ong, Isaac and Almahairi, Amjad and Wu, Vincent and Chiang, Wei-Lin and Wu, Tianhao and Gonzalez, Joseph E. and Kadous, M. Waleed and Stoica, Ion},
  journal = {arXiv preprint arXiv:2406.18665},
  year    = {2024},
  url     = {https://arxiv.org/abs/2406.18665}
}

@article{strati2024dejavu,
  title   = {D{\'e}j{\`a}Vu: {KV}-cache Streaming for Fast, Fault-tolerant Generative {LLM} Serving},
  author  = {Strati, Foteini and McAllister, Sara and Phanishayee, Amar and Tarnawski, Jakub and Klimovic, Ana},
  journal = {arXiv preprint arXiv:2403.01876},
  year    = {2024},
  url     = {https://arxiv.org/abs/2403.01876}
}
}

\end{document}